\relax
\documentclass[letterpaper]{article}
\usepackage{arxiv}
\usepackage{times}
\usepackage{helvet}
\usepackage{courier}
\usepackage[hidelinks]{hyperref}       % hyperlinks
\usepackage{graphicx}
\usepackage{graphicx}
\usepackage[ruled,vlined]{algorithm2e}

\newcommand\blfootnote[1]{%
  \begingroup
  \renewcommand\thefootnote{}\footnote{#1}%
  \addtocounter{footnote}{-1}%
  \endgroup
}

\usepackage{booktabs}       % professional-quality tables
\usepackage{amsfonts}       % blackboard math symbols
\usepackage{nicefrac}       % compact symbols for 1/2, etc.
\usepackage{amsmath}
\usepackage{enumitem}

\begin{document}
\title{Data Pruning: Redundant, Problematic, and Interdependent Samples}

\author{
  Leon Freese$^{1, 2}$, Marthinus W. Theunissen$^{1, 2, 3}$\\ \\
    $^{1}$Faculty of Engineering, North-West University, South Africa\\
    $^{2}$Centre for Artificial Intelligence Research (CAIR), South Africa\\ $^{3}$National Institute for Theoretical and Computational Sciences (NITheCS), South Africa\\
  \{leonfreese5, tiantheunissen\}@gmail.com \\
}

\maketitle

\blfootnote{This work is a preprint of a published paper by the same name~\cite{freese2025data}. The authenticated version is available online at \url{https://link.springer.com/chapter/10.1007/978-3-032-11733-5_12}.}

\begin{abstract}
The performance of deep learning models is affected by not only data quantity but also data quality. Data pruning is a process by which practitioners can reduce the size of a dataset by only keeping the most important training data points, thereby achieving similar test set performance. We empirically investigate two popular data pruning methods under noisy and noiseless conditions and show that these methods fail in the presence of significant label noise. We highlight that the success of data pruning is distinctly affected by three factors: redundancy in the dataset, the presence of problematic samples, and interdependence between samples. We perform a detailed investigation on commonly used benchmark classification datasets and neural network architectures. We find that our observations are consistent across data distributions and training protocols.
\end{abstract}

% keywords can be removed

\section{Introduction}

% \begin{itemize}
%     \item ANNs are cool but dependent on data, yada yada
%     \item Data pruning is often used for...to great sucess...it's important, yada yada
%     \item There is a related subfield or research about detecting problematic samples...yada yada
%     \item A common approach to evaluating the success of such pruning...yada yada
% \end{itemize}

Deep neural networks have garnered significant attention due to their remarkable performance on large-scale datasets. However, their effectiveness fundamentally depends on the quality and quantity of training data~\cite{zhang2016understanding}. With the growing size and complexity of datasets, it has become increasingly evident that not all samples contribute equally to model performance. Datasets can contain statistical outliers~\cite{sehwag2021ssd}, mislabelled samples~\cite{northcutt2021pervasive}, out of distribution instances~\cite{lee2018simple}, redundancies~\cite{paul2021deep}, imbalances~\cite{ren2018learning}, or even \textit{adversarial} examples~\cite{goodfellow2014explaining}. Defining and characterizing how each of these issues affect a model's ability to perform constitutes a wide variety of ongoing investigations in the literature. 
% Some samples may be redundant~\cite{toneva2018an}, slowing the learning process, while others may be problematic~\cite{paul2021deep}, leading to poorer generalization.

Data pruning is the process of removing samples from a train set in a way that allows the model to maintain or improve performance. In addition to possible gains in computational efficiency and robustness, it can also be a useful tool for identifying and analysing training samples to understand their impact on generalization. Data pruning experiments typically involve three steps: 1) ranking training samples based on a \textit{score}, 2) removing a subset of increasing size based on said ranking, and 3) retraining the model on the reduced dataset to evaluate its performance on a test set. This approach has proven effective in variety of investigations. See Section~\ref{sec: related work} for related work. 
% These include mitigating the effects of neural scaling laws, where data pruning can reduce the power-law relationship between test loss and training set size~\cite{sorscher2022beyond}.
% Scoring methods for data pruning typically assign values to samples based on criteria such as their influence on loss~\cite{koh2017understanding}, memorization dynamics~\cite{zhang2020neural}, or gradient behavior~\cite{paul2021deep}. 
% Beyond their direct application in combating scaling laws, these methods enable researchers to study the characteristics and dynamics of training samples in greater detail.

% A related area of research focuses on detecting problematic samples within training datasets, such as mislabeled instances~\cite{northcutt2021confidentlearning}, outliers~\cite{lee2018simple}, or inherently ambiguous or noisy examples. Characterizing these samples provides valuable insights into model behavior, data quality, and the distributional properties of the dataset. Such analyses contribute to a deeper understanding of the open research question of generalization in deep learning.

In this work we contribute the following:
\begin{itemize}
    \item A thorough investigation of the typical data pruning approach.
    \item We show that the order in which samples should be pruned is not as clear as often motivated in the literature since what constitutes ``important'' samples is dependent on confounding factors like redundancy and contradictory samples.
    \item We highlight the large degree of influence the presence of some samples might have on the contribution (to model performance) of others.
\end{itemize}

\noindent
In addition to Section~\ref{sec: related work} summarizing works related to data pruning, Section~\ref{sec: methods} describes our experimental methods in detail. In Section~\ref{sec: results} we present our experimental results along with discussions of our findings. In the final section we conclude with a summarization of our findings.

\section{Related work}
\label{sec: related work}

% \begin{itemize}
%     \item Where is data pruning used?
%     \item What main approaches exist?
%     \item How are these approaches compared in the literature?
%     \item How do these works relate to ours? What gap are we filling?
% \end{itemize}

% The main aspect that varies between different approaches to data pruning is the \textit{score} used to rank the train set samples. Scoring methods for data pruning typically assign values to samples based on criteria such as their influence on loss~\cite{koh2017understanding}, memorization dynamics~\cite{zhang2020neural}, or gradient behavior~\cite{paul2021deep}. 

% mitigating the effects of neural scaling laws, where data pruning can reduce the power-law relationship between test loss and training set size~\cite{sorscher2022beyond}.

Most data pruning experiments are done in the context of \textbf{data valuation}; that is the task of assigning value to specific samples in the dataset. This value\footnote{This is often referred to as a \textit{score}; a convention we will be continuing in this work.} usually refers to how much each sample influences or contributes to the model's overall level of performance~\cite{paul2021deep,sorscher2022beyond,yoon2020data,nohyun2022data}. These \textit{scores} might be useful for determining how much individuals should be compensated for their data~\cite{ghorbani2019data}, but they can also be useful for domain adaptation, corrupted sample discovery, and robust learning~\cite{yoon2020data}.

Ghorbani and Zou~\cite{ghorbani2019data} develop a scoring method based on Shapley values. Using a basic data pruning experiment they show that, if samples are pruned in descending order, there is a more pronounced drop in performance than if pruned in a random order or based on a classic baseline scoring method. They also compare performance if samples are pruned in ascending order. In this case the performance tends to improve with each increment. This indicates that their method scores samples in a way that correlates with their contribution to overall model performance. This approach was later generalized by Kwon et al.~\cite{kwon2021beta} to similar effect.

Yoon et al.~\cite{yoon2020data} propose an alternative scoring method using a reinforcement learning approach. Using a similar data pruning experiment to Ghorbani and Zou, they show that their scoring method outperforms that of Ghorbani and Zou and other baselines on common benchmarks and label corrupted alternatives.

Paul et al.~\cite{paul2021deep} propose two scoring methods based on gradient- and error dynamics measured early in training. Their data pruning experiments show that when samples are pruned in ascending order, the model maintains good performance at much lower train set sizes than when samples are pruned randomly. Interestingly, they find that label noise tends to increase their scores, but don't show the effect on an actual pruning experiment. More recently, Ki et al.~\cite{nohyun2022data} proposed a training-free alternative that performs competitively without having to train a model in order to generate scores.

Data pruning experiments have also been used to evaluate \textbf{label error detectors} specifically. For example, Pleiss et al.~\cite{pleiss2020identifying} propose a method for detecting problematic (i.e. mislabelled) samples using an \textit{Area Under the Margin (AUM)} statistic. They use a basic data pruning experiment to show that, when samples are ranked according to this statistic, the model performance improves above an alternative random ranking.

Northcutt et al.~\cite{northcutt2021confidentlearning} propose a method of detecting label errors based on the class output probabilities of a trained model. % Through a variation of data pruning experiments they show that, when pruning progressively larger portions of the samples detected as label errors, model performance increases more than when pruning random samples.
They evaluate this approach through a variation of data pruning experiments. Their results show that pruning progressively larger portions of the samples detected as label errors improves model performance more than pruning random samples.

\textbf{Other works} do not propose data valuation approaches or label error detectors directly, but use data pruning experiments to either probe the inner workings of machine learning models or use it as a tool to analyse expected model performance under different conditions.

Toneva et al.~\cite{toneva2018an} explore the nature of sample forgetting in deep neural networks. By means of data pruning experiments, they show that if samples are pruned in ascending order of the number of forgetting events, the model is able to maintain full-set performance at smaller train set sizes than if the samples are pruned randomly. This suggests that samples with more forgetting events are more important for model performance. Toneva et al. also mention that label noise tends to increase their scores without demonstrating the implications on pruning experiments.

Looking into how deep neural networks treat individual samples during training, Jiang et al.~\cite{jiang2020characterizing} propose a theoretical score that quantifies the expected accuracy on a held-out sample when train sets of varying sizes are drawn from the data distribution. They show that an approximated version of their score can be used as a successful scoring method for data pruning when compared to a random baseline.

Recently, Sorscher et al.~\cite{sorscher2022beyond} argue that with powerful data pruning metrics we are able to break beyond the power law scaling~\cite{hoffmann2022training} that is often observed when comparing performance while varying model- or dataset size. %Through an extensive set of data pruning experiments (including ten data pruning approaches) they find that most approaches fail to scale to large-scale datasets like ImageNet~\cite{deng2009imagenet} and the most powerful approaches are often computationally more expensive. They also develop a theoretical framework to characterize data pruning under simplified conditions (perceptron learning in the student-teacher setting), suggesting that if the initial dataset is large, one should prune ``easy'' samples first, and if the initial dataset is small one should prune ``hard'' samples first. 
They support this argument through an extensive set of data pruning experiments, which include ten different pruning approaches. Their results show that most methods fail to scale to large datasets such as ImageNet~\cite{deng2009imagenet}. Moreover, the most effective approaches are often computationally expensive. In addition, they develop a theoretical framework to characterize data pruning under simplified conditions (perceptron learning in the student–teacher setting). This framework suggests that when the initial dataset is large, one should prune ``easy'' samples first, whereas when the dataset is small, one should prune ``hard'' samples first. Here easy would correspond to large margins in the teacher and hard samples would correspond to small margins.

We note that none of the works mentioned above pruned the train set to near $100 \%$ while comparing with a baseline. We also note that few of the works perform data pruning experiments with explicit label noise. The exceptions being Yoon et al~\cite{yoon2020data}, and the two label error detectors Pleiss et al~\cite{pleiss2020identifying}. and Northcutt et al.~\cite{northcutt2021confidentlearning}. Finally, we note the works by Paul et al.~\cite{paul2021deep}, Ki et al.~\cite{nohyun2022data}, and Toneva et al.~\cite{toneva2018an}, all mention that their scores tend to be higher for label noise. However, they maintain the argument that samples with higher scores tend to be more important for model performance (hence why they prune in ascending order of score).

These points prompted us to take a closer look at how two popular scoring methods~\cite{paul2021deep,toneva2018an} perform as data pruning approaches with and without label noise across the full range of the train set that can be pruned.

\section{Methods}
\label{sec: methods}

% \subsection{Data pruning}

% \begin{itemize}
%     \item Define our evaluation framework precisely.
%     \item Disclose any part of our methods that will remain constant throughout the paper here already.
%     \item Provide neat algorithm.
%     \item Add short discussion of limitations.
% \end{itemize}

\subsubsection{Problem formulation} In a multiclass classification problem, let the input be $\mathbf{x} \in X \subseteq \mathbb{R}^{d}$ and the label space $Y = \{1, 2, \dots, c\}$. The labelled dataset is defined as $D = \{(\mathbf{x}_i, y_i)\}_{i=1}^n$, where each pair $(\mathbf{x}_i, y_i)$ is called a \emph{sample}. $D$ is divided into two non-overlapping subsets: $D_{\text{train}} \cup D_{\text{test}}$. We assume that $D_{\text{test}}$ is entirely drawn \emph{independent and identically distributed (i.i.d.)} from some target distribution over $X \times Y$. We do not make the same assumption for $D_{\text{train}}$ due to potential noise.

% In this work, we define two types of training samples based on their impact on test set performance: \emph{redundant samples} and \emph{problematic samples}.

% \begin{definition}
%     Redundant sample. A training sample $(\mathbf{x}_i, y_i)$ is considered redundant if including it in $D_{\text{train}}$ results in the same performance metric evaluated on $D_{\text{test}}$ as if it were excluded.
% \end{definition}

% \begin{definition}
%     Problematic sample. A training sample $(\mathbf{x}_i, y_i)$ is considered problematic if including it in $D_{\text{train}}$ results in a worse performance metric evaluated on $D_{\text{test}}$ than if it were excluded. The greater the degradation in the performance metric, the more problematic the sample.
% \end{definition}

\subsubsection{Data pruning} As is convention in the literature, we use a framework containing a model architecture $\mathcal{M}$, a training process $\mathcal{T}$, a scoring method $\mathcal{S}$, and a performance measure $\mathcal{P}$. For $D_{\text{train}}$, the training process is defined as $\mathcal{T}(D_{\text{train}}, \mathcal{M}) \to \mathcal{M}_{\text{train}}$
% \begin{equation}
%     T: D_{\text{train}} \times \mathcal{M} \to \mathcal{M}_{\text{train}} \times \mathbb{R}, \quad T(D_{\text{train}}, \mathcal{M}) = (\mathcal{M}_{\text{train}}, p_{\text{train}})
% \end{equation}
% \begin{equation}
%     \mathcal{T}(D_{\text{train}}, \mathcal{M}) \to \mathcal{M}_{\text{train}}, p_{\text{train}}
% \end{equation}
where $\mathcal{M}_{\text{train}}$ is the model trained on $D_{\text{train}}$.
Similarly, for $D_{\text{train}}$, the performance metric is defined as $\mathcal{P}(D_{\text{train}}, \mathcal{M}_{\text{train}}) \to p_{\text{train}}$ and represents the overall model performance on the given dataset.
The scoring method is defined as $\mathcal{S}(\mathcal{T}) \to \{s_i\}_{i=1}^n$
% \begin{equation}
%     S: D_{\text{train}} \times \mathcal{M} \times T \to \mathbb{R}^n, \quad S(D_{\text{train}}, \mathcal{M}, T) = \{s_i\}_{i=1}^n
% \end{equation}
% \begin{equation}
%     \mathcal{S}(D_{\text{train}}, \mathcal{M}, \mathcal{T}) \to \{s_i\}_{i=1}^n
% \end{equation}
where each $s_i \in \mathbb{R}$ is the score assigned to the training sample $(\mathbf{x}_i, y_i) \in D_{\text{train}}$. The data pruning evaluation framework outlined in Algorithm~\ref{alg: psi curve} produces a $\psi$-curve that represents the expected model performance, on a held-out test set, as an increasing portion of the train set is pruned.

If $\mathcal{S}$ produces scores that correlate with how much each training sample contributes to the overall model performance, the $\psi$-curve should either maintain stable performance or have an initial upwards trend as more samples are being pruned. This is because Algorithm~\ref{alg: psi curve} ranks the train samples in ascending order, meaning the least likely to contribute to model performance should be pruned first. In general, an unbiased ranking of the train set should correspond to a gradual downward slope in the $\psi$-curve since model performance is known to be dependent on the train set size.

By using different approaches to scoring and ranking the train set we can compare data pruning metrics and investigate the role of specific samples in the train set and how they contribute to model performance.

\begin{algorithm}[h]
\KwIn{$D_{\text{train}}$, $D_{\text{test}}$, $\mathcal{M}$, $\mathcal{T}$, $\mathcal{S}$, and $\mathcal{P}$ as defined above. As well as $r$, a resolution term for which $1 \leq r \leq |D_{\text{train}}|$ and $|D_{\text{train}}|\ mod\ r = 0$.}
\KwOut{A $\psi$-curve of length $|D_{\text{train}}| / r$.}

% \nl $\mathcal{S}_{\text{train}} \leftarrow \Delta(\mathbb{D}_{\text{train}}, \mathbf{\Theta}, \Lambda)$ \\

% \nl Define $\mathbb{Z} = \{z\}_{i=0}^{|\mathbb{D}_{\text{train}}|}$ such that $\mathbb{S}_{\text{train}}[z_{i}] \leq \mathbb{S}_{\text{train}}[z_{i+1}]: \{i=0, ... , |\mathbb{D}_{\text{train}}| - 1\}$ \\
\nl $\mathbb{Z} = argsort(\mathcal{S}(\mathcal{T}(D_{\text{train}}, \mathcal{M})))$ \\

\nl $\psi \leftarrow ()$ \\

\nl \For{$i$ in $(0, ... , |D_{\text{train}}| / r)$}{

\nl $D_{\text{trunc}} = D_{\text{train}} - D_{\text{train}}^{\mathbb{Z}^{\{0, ..., r \times i\}}}$ \\

\nl $\mathcal{M}_{trunc} = \mathcal{T}(D_{\text{trunc}}, \mathcal{M})$ \\

\nl $p_{\text{test}} \leftarrow \mathcal{P}(D_{\text{test}}, \mathcal{M}_{\text{trunc}})$

\nl $\psi \leftarrow \psi + p_{test}$

}
\nl return $\psi$
\caption{{\bf Generating a $\Psi$-curve} \label{alg: psi curve}}
\end{algorithm}

\subsection{Scoring methods}
\label{sec: scoring methods}

% \begin{itemize}
%     \item Provide exact definitions of the scoring methods we will employ in this paper (let's call them scoring methods instead of detectors, what do you think)?
%     \item Use clear math notation.
%     \item Motivation why we chose them.
% \end{itemize}

% To identify redundant and problematic samples within the training set $D_{\text{train}}$, as defined in our framework, we adopt two established scoring methods from the literature: the sample forgetting score~\cite{toneva2018an}, which leverages the forgetting behavior of samples during training, and the error L2-norm (EL2N) score~\cite{paul2021deep}, which quantifies early-training prediction errors. These methods assign scores $\{s_i\}_{i=1}^n$ to each training sample $(\mathbf{x}_i, y_i) \in D_{\text{train}}$, enabling the detection of samples that minimally or detrimentally impact model performance on $D_{\text{test}}$.

In this work we make use of two established scoring methods from the literature: the \textit{sample forgetting score} of Toneva et al.~\cite{toneva2018an} and the \textit{error $L_2$-norm (EL2N) score} from Paul et al.~\cite{paul2021deep}. Both of these methods are well-cited and are often compared with by other works~\cite{jiang2020characterizing,nohyun2022data,sorscher2022beyond}. Additionally, they have been shown to perform at a similar level to alternatives at scale in the extensive comparison by Sorscher et al.~\cite{sorscher2022beyond}, even producing similar rankings to some alternatives.

%\subsubsection{Sample Forgetting Score} Toneva \emph{et al.}~\cite{toneva2018an} propose a scoring method based on the forgetting behaviour of neural networks, inspired by catastrophic forgetting, where a model loses previously learned information when trained on new tasks. 
\subsubsection{Sample Forgetting Score} Toneva et al.~\cite{toneva2018an} propose a scoring method based on the forgetting behaviour of neural networks. The method is inspired by catastrophic forgetting, where a model loses previously learned information when trained on new tasks. In the current context, each iteration in the stochastic gradient descent process is treated as a \emph{mini-task}. A training sample is considered \emph{forgotten} at iteration $t$ if the model correctly classifies it at iteration $t-1$, but misclassifies it at $t$. Here, iterations correspond to consecutive mini-batches containing the sample. To simplify computation, Toneva et al.~\cite{toneva2018an} use epochs as a lower bound for iterations. The forgetting events for each sample are summed across the full training cycle to compute a cumulative \emph{forgetting score} $s_i$. Samples that are learned once and never forgotten are termed \emph{unforgettable} and receive a score of zero, while samples that are never correctly classified receive an \emph{infinite} forgetting score.

In other words, a sample $(\mathbf{x}_i, y_i) \in D_{\text{train}}$ observed over $e$ iterations during the training process $\mathcal{T}$ is given a \textit{forgetting score} of:
\begin{equation}
    s_{i} = \sum_{t = 1}^{e} \mathbf{1}_{\{f_{t}(\mathbf{x}_i) \neq y_i \& f_{t-1}(\mathbf{x}_i) = y_i\}}
    \label{eq: forget}
\end{equation}
where $f_{t}(\mathbf{x}_i)$ is the model's class prediction for input $\mathbf{x}_i$ at iteration $t$.

\subsubsection{Error $L_{2}$-Norm (EL2N) Score} Paul et al.~\cite{paul2021deep} introduce a score, which leverages early-training model behaviour. 
% For a training sample $(\mathbf{x}_i, y_i) \in D_{\text{train}}$, let $\mathbf{p}(\boldsymbol{\phi}_t, \mathbf{x}_i)$ denote the model’s output probability vector at epoch $t$, where $\boldsymbol{\phi}_t$ represents the model parameters after $t$ epochs of SGD training, and let $\mathbf{y}_i$ be the one-hot-encoded label of the sample. The EL2N score is defined as:
% \begin{equation}
%     s_i = \mathbb{E}_{\boldsymbol{\phi}_t} \|\mathbf{p}(\boldsymbol{\phi}_t, \mathbf{x}_i) - \mathbf{y}_i\|_2
% \end{equation}
This score is simply the average Euclidean norm of the error vector of the model's prediction with respect to the current sample at a relatively early stage in training.
More specifically, for a training sample $(\mathbf{x}_i, y_i) \in D_{\text{train}}$, the \textit{EL2N score} is defined as:
\begin{equation}
    s_{i} = \mathbb{E}\|\mathbf{f}_{t}(\mathbf{x}_{i}) - \mathbf{y}_{i} \|_{2}
    \label{eq: el2n}
\end{equation}
where $\mathbf{f}_{t}(\mathbf{x}_{i})$ denote the model’s output probability vector at iteration $t$ for input $\mathbf{x}_{i}$ and $\mathbf{y}_{i}$ is a one-hot-encoding of $y_{i}$. The expected value is estimated by averaging across multiple model initializations. 
% High EL2N scores indicate samples that are misclassified early in training, potentially marking them as problematic, while low scores suggest samples that are easily learned, which may be redundant.

Note that both the \textit{sample forgetting score} and the \textit{EL2N score} are grounded in intuitions about the ease with which a sample is fitted by the model. The former is based on how many times a model unlearns the features required to correctly classify it and the latter estimates how wrong the model is on producing the correct output value when training iterations are limited. This aligns with the views of Sorscher et al.~\cite{sorscher2022beyond} about the best strategy for pruning being to prune easy samples first when the data is abundant. Both Toneva et al.~\cite{toneva2018an} and Paul et al.~\cite{paul2021deep} prune in ascending order of their scores.

\subsection{Experimental setup}

% \begin{itemize}
%     \item What models are we training?
%     \item What datasets?
%     \item What was our training protocol?
%     \item Any thing related to the ordering, scoring, seeds, and retraining should be disclosed by this point.
% \end{itemize}

\subsubsection{Datasets} We conduct experiments on three multiclass classification datasets: a synthetic dataset and two benchmark datasets, MNIST~\cite{deng2012mnist} and CIFAR-10~\cite{krizhevsky2009learning}. For the synthetic dataset, we generate samples $(\mathbf{x}_i, y_i)$ such that $\mathbf{x}_i \in \mathbb{R}^{100}$ is uniformly sampled from one of ten class-specific isotropic Gaussian distributions $\{\mathcal{N}(\boldsymbol{\mu}_j, \sigma^{2}_{j}\mathbf{I)}\}_{j=1}^{10}$, where $\sigma_{j} \sim \mathcal{U}(1,5)$, and all $\boldsymbol{\mu}_j$ are separated by a Euclidean distance of seven. Our train- and test set sizes are reported in Table~\ref{tab:config}. The MNIST and CIFAR-10 train sets are randomly sampled from the full sets and the test sets are the predefined evaluation datasets. For experiments involving label noise, we apply it only to the train set. The noise is applied symmetrically by randomly replacing the labels of a subset of the training samples with other labels randomly chosen from the label space.

\subsubsection{Models} All models are trained using the Adam~\cite{adam2014method} optimizer with a batch size and learning rate as specified in Table~\ref{tab:config}. Batch normalization~\cite{ioffe2015batch} layers are applied after each hidden layer, before a ReLU activation function. For the synthetic and MNIST datasets we train Multilayer Perceptron (MLP) models, and for CIFAR-10 we train VGG-16 models. The MLP models consists of two hidden layers, each with $512$ nodes. For the VGG-16 models, the last 3 layers of the model are three fully connected layers with $4,096$, $4,096$ and $1,000$ nodes, as suggested by Simonyan et al.~\cite{simonyan2015very}.

\subsubsection{Scoring} We compute the scores $\{s_i\}_{i=1}^n$ using each of the two scoring methods by averaging the results over ten randomly initialized models. To count the forgetting events as in Eq~\ref{eq: forget}, each model is trained for $200$ epochs. The error vectors for Eq~\ref{eq: el2n} are measured after $20$ epochs of training. These design choices were made to closely resemble the setups in the works by Toneva et al.~\cite{toneva2018an} and Paul et al.~\cite{paul2021deep}.
% We then evaluate these scores using three evaluation models, each initialized with random seeds distinct from those of the detector models. For each scoring method, we report the mean performance across the three evaluation runs, along with the standard deviation, to assess the robustness of the scores. The architecture and hyperparameters of both detector and evaluation models are held constant for each dataset, respectively.

\subsubsection{Pruning} After calculating a score for each training sample, we use Algorithm~\ref{alg: psi curve} to generate a $\psi$-curve. In practice, at each point along the curve, we evaluate the performance over three random initializations of the model architecture. Importantly, we use the same model architecture $\mathcal{M}$ and training process $\mathcal{T}$ as the ones used to calculate the scores. The resolution parameter $r$ is set as shown in Table~\ref{tab:config}. In addition to the two scoring methods, we perform data pruning experiments using a random baseline, where the sample order ($\mathbb{Z}$ in Algorithm~\ref{alg: psi curve}) is replaced with a random permutation.

% We use the Adam optimizer for all models, with a batch sizes specified in Table~\ref{tab:config}. ReLU is used as the activation function throughout. Batch normalization layers is applied after each layer. The MLP models consists of 2 hidden layers, each with 512 nodes. For the VGG-16 model, the last 3 layers of the model are 3 fully connected layers with 4,096, 4,096 and 1,000 nodes per layer respectively, the same as proposed by Simonyan \emph{et al.}~\cite{simonyan2015very}, each followed by ReLU activation.

\section{Results}
\label{sec: results}

\begin{figure}[t]
    \centering
    \includegraphics[width=0.49\linewidth]{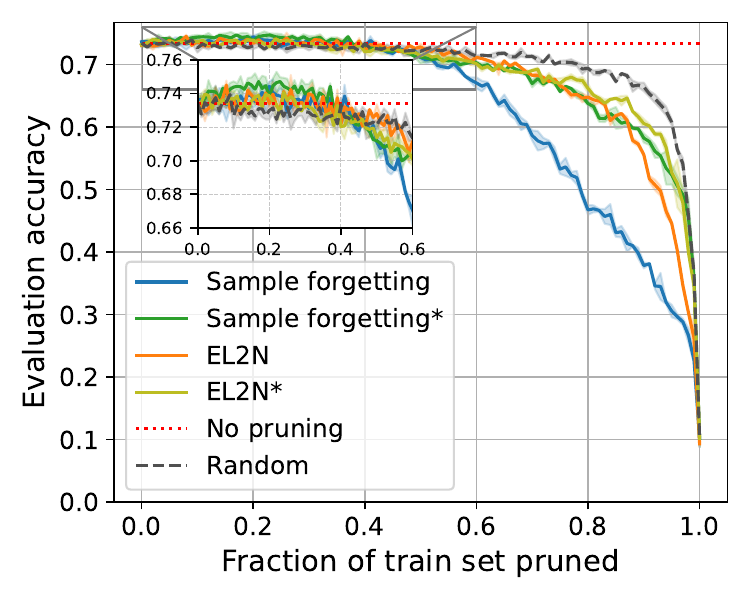}
    \includegraphics[width=0.49\linewidth]{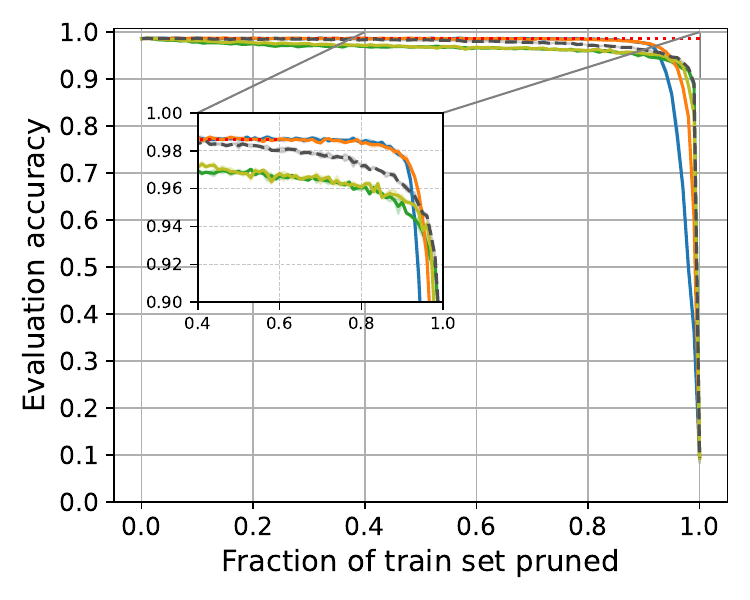}
    \includegraphics[width=0.49\linewidth]{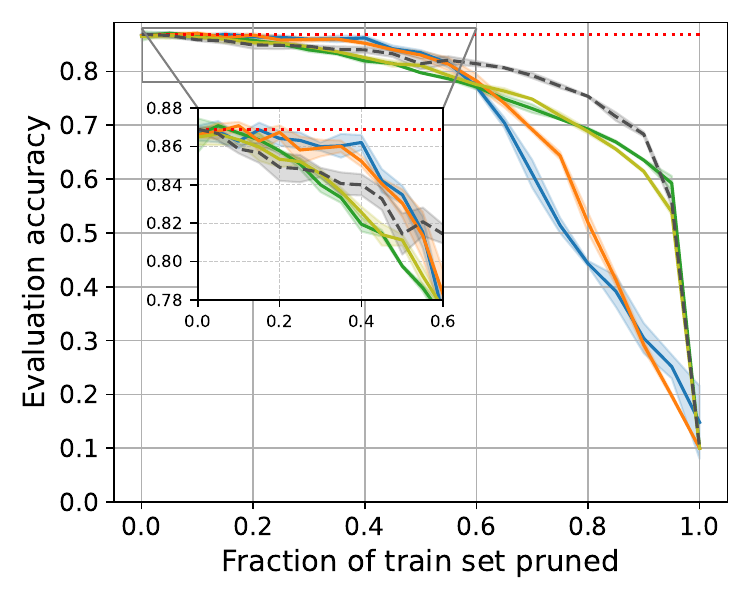}
    \caption{Model performance as a function of pruning an increasing portion of the train set for synthetic (left-top), MNIST (right-top), and CIFAR-10 (bottom) datasets. Samples are pruned in ascending order, with rankings provided by the two scoring methods in Section \ref{sec: scoring methods}, or by a random baseline. The methods with a $*$ in the title have their ranking reversed. The legend in the first plot is applicable to all three plots.}
    \label{fig:clean_synth}
\end{figure}

From Fig.~\ref{fig:clean_synth} we note that we can prune roughly $40\%$, $90\%$, and $40\%$ of the synthetic, MNIST, and CIFAR-10 train set, respectively, with negligible reductions in the expected model performance. This aligns with others' work where it is found that performance can be maintained while pruning $80\%$~\cite{toneva2018an} of the full MNIST train set and $30\%$~\cite{toneva2018an} to $50\%$~\cite{paul2021deep} of the full CIFAR-10 train set. There is clearly a high level of redundancy in these datasets.

% The results in Fig.~\ref{fig:clean_synth} align with others' work where it is found that performance can be maintained while pruning $80\%$~\cite{toneva2018an} of the full MNIST train set and $30\%$~\cite{toneva2018an} to $50\%$~\cite{paul2021deep} of the full CIFAR-10 train set. There is clearly a high level of redundancy in these datasets.

Here we point out that when considering the full range of the horizontal axes in Fig~\ref{fig:clean_synth}, the improvements of the two pruning approaches (blue and orange) over the random baseline (gray) appear less significant than when only focussing on the left side of the plots. At the previously mentioned thresholds, we see a gain of approximately $2\%$ accuracy for CIFAR-10 and even less for MNIST and the synthetic dataset.

Additionally, we see that the random baseline maintains higher model performance at extremely low train set sizes. If the general intuition is that the samples pruned last are \textit{important} for generalization, why would the random baseline models outperform the two scoring methods so consistently in the most extreme cases (i.e. when only keeping the most important samples)? We hypothesize that the reason for this observation is that redundancy can be pruned, but not removed entirely. Samples that might be redundant at large train sizes can still be important for generalization at small train sizes. The random baseline prunes samples arbitrarily, resulting in samples that were redundant at large train sizes still being present at small train sizes. In contrast, the scoring methods excessively prune redundant samples at large train sizes.

To support this hypothesis we reverse the order of the rankings of the two scoring methods and repeat the pruning experiments. This is indicated by the $*$'d (green and olive) curves. In this case, when an extreme portion of the train set is pruned (only keeping the ``least important'' samples), the models perform better than when keeping the ``most important'' samples. This contradiction somewhat aligns with the framework of Sorscher et al.~\cite{sorscher2022beyond} stating that one should prune easy samples when data is abundant and prune hard samples when data is scarce. However, the fact that the models pruned by the random baseline outperforms all the scoring methods in the most extreme cases of data pruning suggests that one requires a mixture of easy and hard samples when pruning to such an extent.

The main takeaway is that when the train set contains substantial redundancy, many samples can be removed (even with random pruning) with little impact on model performance. However, excessive pruning using more powerful scoring methods can cause disproportionate performance drops, as redundant samples are not necessarily inherently unimportant.

\begin{figure}[h]
    \centering
    \includegraphics[width=0.49\linewidth]{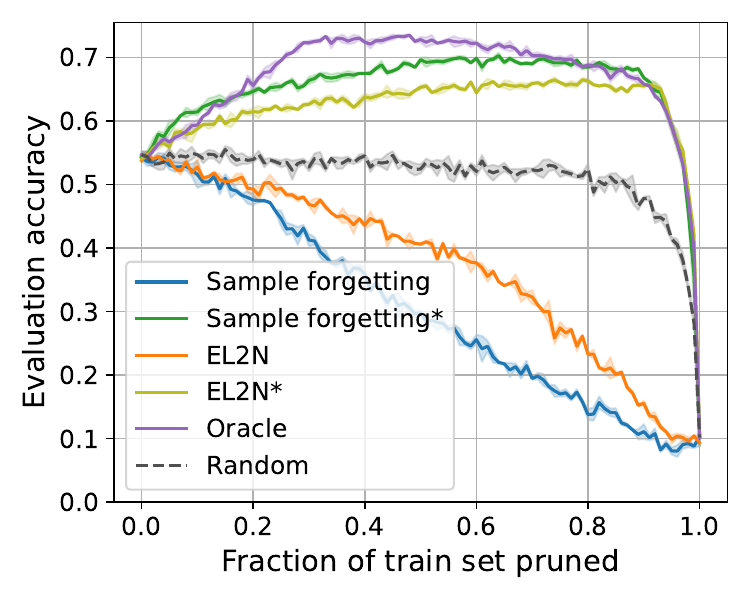}
    \includegraphics[width=0.49\linewidth]{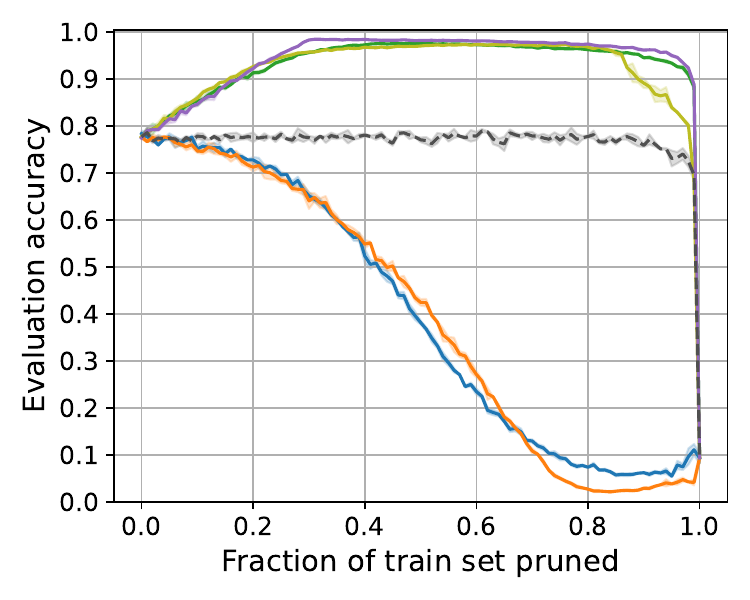}
    \includegraphics[width=0.49\linewidth]{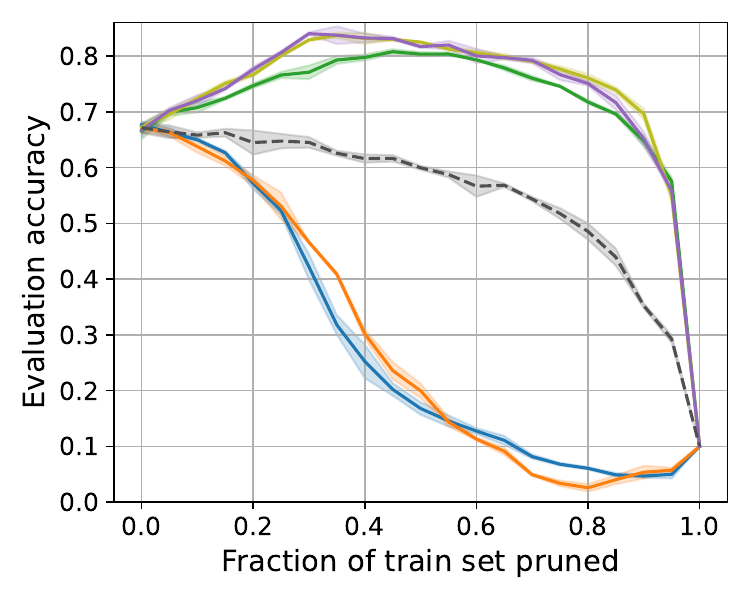}
    \caption{Model performance as a function of pruning an increasing portion of the train set with $30\%$ label noise for synthetic (left-top), MNIST (right-top), and CIFAR-10 (bottom) datasets. Samples are pruned in ascending order, with rankings provided by the two scoring methods in Section \ref{sec: scoring methods}. The methods with a $*$ in the title have their ranking reversed. The \textit{Random} method provides a random baseline. The \textit{Oracle} method first prunes label corrupted samples in a random order, and then prunes the rest of the train set in a random order. The legend in the first plot is applicable to all three plots.}
    \label{fig:noisy_synth_asc}
\end{figure}

In Fig.~\ref{fig:noisy_synth_asc} we see that when using the original \textit{sample forgetting} (blue) and \textit{EL2N} (orange) scoring methods in the presence of strong label noise, model performance degrades catastrophically during pruning. In the work by Tenova et al.~\cite{toneva2018an} it is mentioned that label corrupted samples tend to have higher forgetting scores. In the work by Paul et al.~\cite{paul2021deep} it is mentioned that training with only high scoring samples in a noisy train set might not be optimal. Building on their observations, we analyse the ranking of samples using \textit{EL2N} and \textit{Sample forgetting} scores for our clean and noisy samples before and after label are corrupted in Fig.~\ref{fig:synth_scores_before_after} in Appendix~\ref{sec: noise scores}. We find that label corrupted samples have a strong tendency to be ranked last (therefore dropped last). We also see that there is a tendency for samples that are never corrupted to maintain their relative position in the ranking after the dataset is corrupted. In contrast corrupted samples that were ranked last before corruption tend to be ranked earlier after corruption. The clarity of these tendencies vary across the three datasets, but it can be noticed consistently.

% \begin{figure}[h]
%     \centering
%     \includegraphics[width=0.49\linewidth]{Figures/synth_multi_30_symm_oracle_forgetting_el2n_random_plot.pdf}
%     \includegraphics[width=0.49\linewidth]{Figures/mnist_30_symm_oracle_forgetting_el2n_random_plot.pdf}
%     \includegraphics[width=0.49\linewidth]{Figures/cifar_30_symm_oracle_forgetting_el2n_random_plot.pdf}
%     \caption{Model performance as a function of pruning an increasing portion of the train set with $30\%$ label noise for synthetic (left-top), MNIST (right-top), and CIFAR-10 (bottom) datasets. Samples are pruned in ascending order, with rankings provided by the two scoring methods in Section \ref{sec: scoring methods}. The methods with a $*$ in the title have their ranking reversed. The \textit{Random} method provides a random baseline. The \textit{Oracle} method first prunes label corrupted samples in a random order, and then prunes the rest of the train set in a random order. The legend in the first plot is applicable to all three plots.}
%     \label{fig:noisy_synth_asc}
% \end{figure} 
% Clearly the samples that have low \textit{sample forgetting} and \textit{EL2N} scores are no longer redundant for model performance. 

% Either the samples that were previously scored low are no longer scored low in the presence of label noise, or they are scored similarly, but they are no longer redundant for model performance. To confirm that it is the latter, we plot the scores of samples 

By considering these observations we find that a simple trick to overcome the drop in performance is to reverse the ranking order before pruning the dataset. This is represented in Fig.~\ref{fig:noisy_synth_asc} with the green and olive curves. Note that when the train set is pruned in descending order we not only see that the model performance is maintained, but improves initially as the label noise is ranked higher, therefore being pruned first. Strikingly, for MNIST and CIFAR-10, the \textit{EL2N score} produces model performances similar to an Oracle ranking which first prunes label noise before pruning clean samples.

\section{Conclusion}
\label{sec: conclusion}

In this work we have performed a controlled set of data pruning experiments to investigate a neglected aspect of data pruning: the influence of label noise on the actual ranking of scoring methods. In the setting that we perform our investigation, we found that:
\begin{enumerate}
    \item When the data set contains a lot of redundancy, the use of \textit{sample forgetting} and \textit{EL2N} scoring methods produces marginal gains over a random baseline under moderate pruning.
    % \item At extreme levels of data pruning, the random baseline produces models that outperform those pruned using the \textit{sample forgetting} and \textit{EL2N} scoring methods, despite these methods being intended to retain the most important samples for model performance.
    \item At extreme levels of data pruning, the random baseline produces models that outperform those pruned with the \textit{sample forgetting} and \textit{EL2N} scoring methods. This is notable because these methods are specifically designed to retain the most important samples for model performance.
    \item Reversing the ranking from \textit{sample forgetting} and \textit{EL2N} scores has little effect under moderate pruning, but paradoxically improves model performance under extreme pruning. This indicates that a sample's score depends on the presence of other samples in the dataset.
    \item In the presence of label noise the \textit{sample forgetting} and \textit{EL2N} scoring methods fail completely, but simply reversing the ranking orders results in drastic improvements to model performance. 
\end{enumerate}

\noindent
Our study has certain limitations. 
% First, regarding the scoring methods, while reversing the removal order effectively mitigates the impact of label noise and improves performance under extreme pruning, it introduces a trade-off under moderate pruning: in such cases, the original methods can still provide modest performance gains over the reversed order. Second, our evaluation framework itself presents a limitation. Like prior work, we remove a fixed number of samples at each pruning step. A more flexible approach would be to adopt a dynamic resolution parameter, allowing the number of samples removed at each step to vary according to sample interdependencies. We leave this extension to future work, as incorporating such dynamics may further improve pruning effectiveness and better reflect the complex interactions between samples.
\begin{itemize}
    \item Whether our findings generalize to larger datasets (e.g. ImageNet) is undetermined.
    \item Our use of artificial label noise might produce an overly optimistic characterization of the expected performance of these pruning experiments. In real-world scenarios, a broader range of problematic samples exists, each with the potential to introduce unforeseen effects on pruning experiments.
    \item Our experiments are also all performed on image classification or image classification adjacent problems. Whether the findings extend to other domains, such as time-series regression, remains uncertain.
\end{itemize}

% While our study is primarily experimental and focused on controlled data pruning scenarios, the findings have clear relevance to real-world machine learning applications. In practice, datasets are often large, redundant, and noisy, and practitioners must make decisions about which samples to prioritize or remove to improve efficiency or reduce training costs. Our results highlight the limitations of current pruning metrics in the presence of label noise and extreme data reduction, providing guidance for selecting or designing novel scoring methods in practical settings.

\noindent
We hope that these observations guide future data pruning metric development as motivated by Sorscher et al.~\cite{sorscher2022beyond} and others. Future work includes incorporating knowledge of sample interdependence into the ranking process, exploring alternative scoring methods and their sensitivity to label noise, and examining the effects of other types of problematic samples.

% All these results provide a few key take-aways:
% \begin{enumerate}
%     \item When your dataset has a lot of redundant samples, you can crudely drop the ones that are most easily fitted by a proxy model and expect to get computational gains.
%     \item if your dataset has a lot of target noise, it is best to drop the samples that the proxy model struggles with the most, up to a point.
%     \item Samples might be i.i.d but their contribution to model performance certainly are not. So you need to be careful about what samples you drop together.
% \end{enumerate}

% Depending on how well we do with 4.3 we can emphasise it more or less.

\subsubsection{Acknowledgment}
This work is based on the research supported in part by the National Research Foundation of South Africa (Grant Reference Numbers: RCDL240215206999).

\bibliographystyle{unsrt}
\bibliography{refs}

\appendix

\section{Appendix}
\label{sec: noise scores}

\begin{table}[ht]
\caption{Dataset sizes and model configurations for the three datasets.}
\label{tab:config}
\centering
\begin{tabular}{|l|l|l|l|l|l|l|}
\hline
Dataset & $|D_{\text{train}}|$ & $|D_{\text{test}}|$ & $r$ & Model & Batch size & Learning rate\\
\hline
Synthetic & 10,000 & 5,000 & 100 & 2-layer MLP & 100 & 0.01\\
MNIST & 55,000 & 10,000 & 550 & 2-layer MLP & 64 & 0.01\\
CIFAR-10 & 45,000 & 10,000 & 2,250 & VGG-16 & 64 & 0.001\\
\hline
\end{tabular}
\end{table}

\begin{figure}[h]
    \centering
    \includegraphics[width=0.45\linewidth]{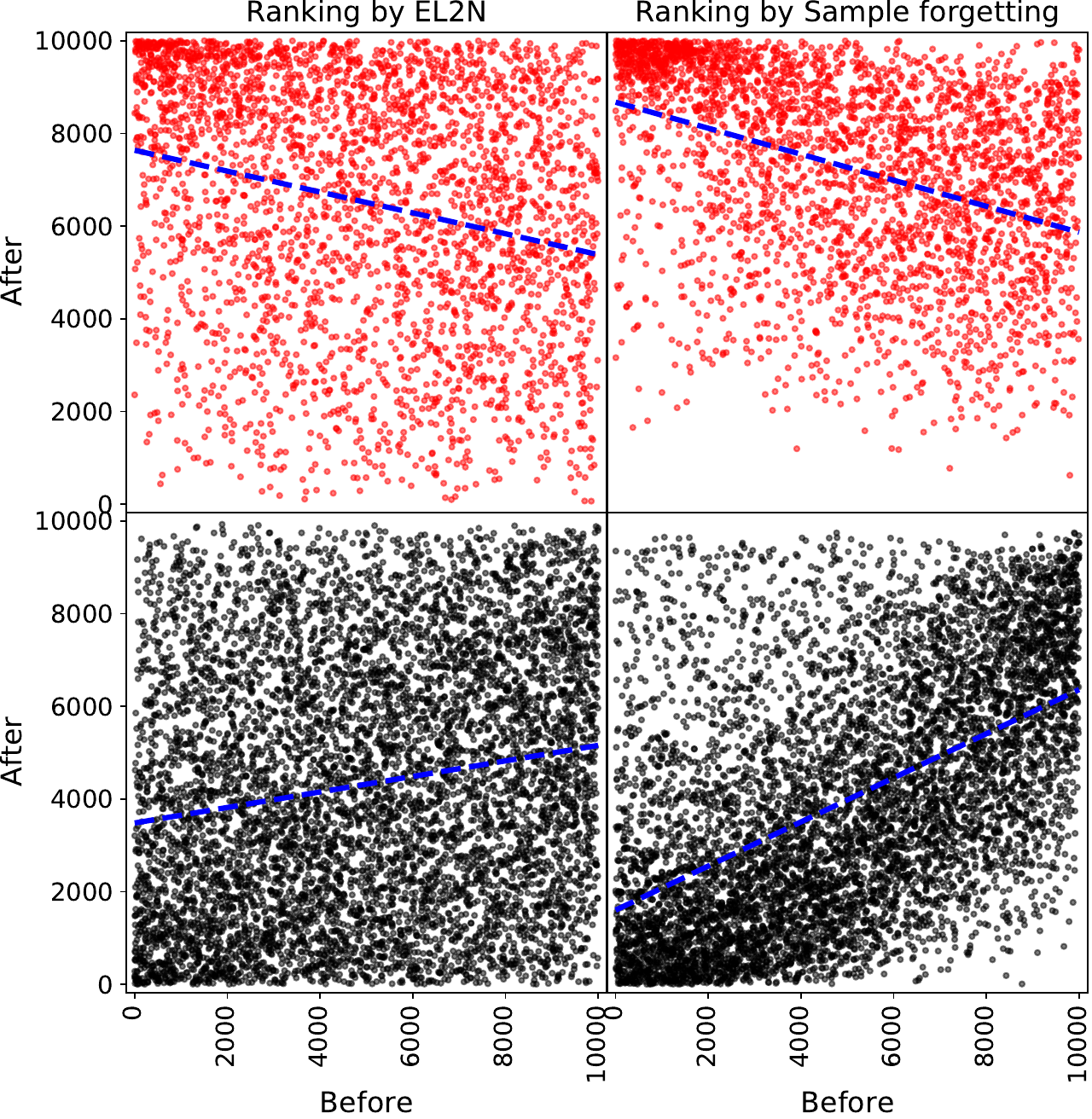}
    \includegraphics[width=0.45\linewidth]{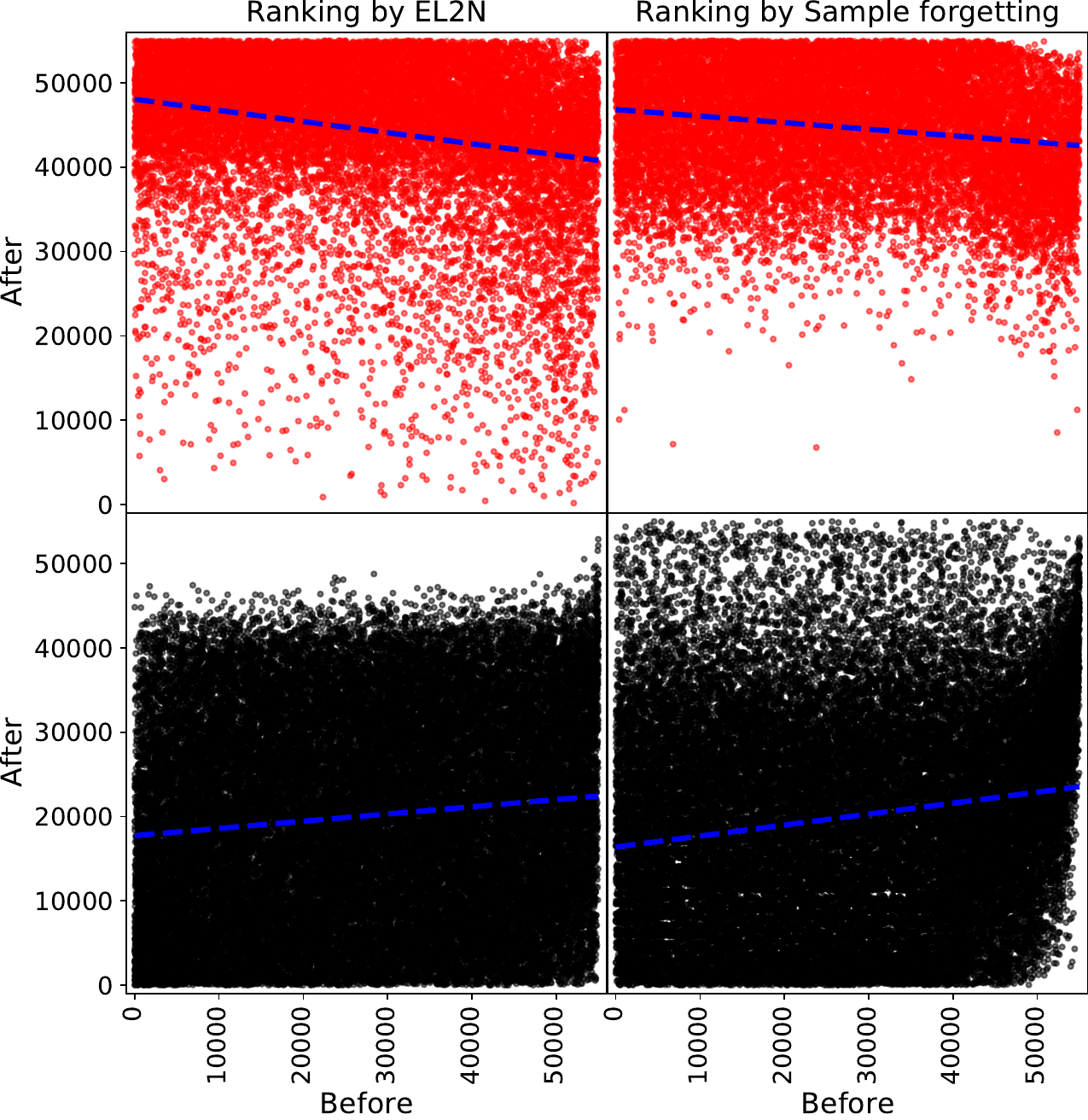}
    \includegraphics[width=0.45\linewidth]{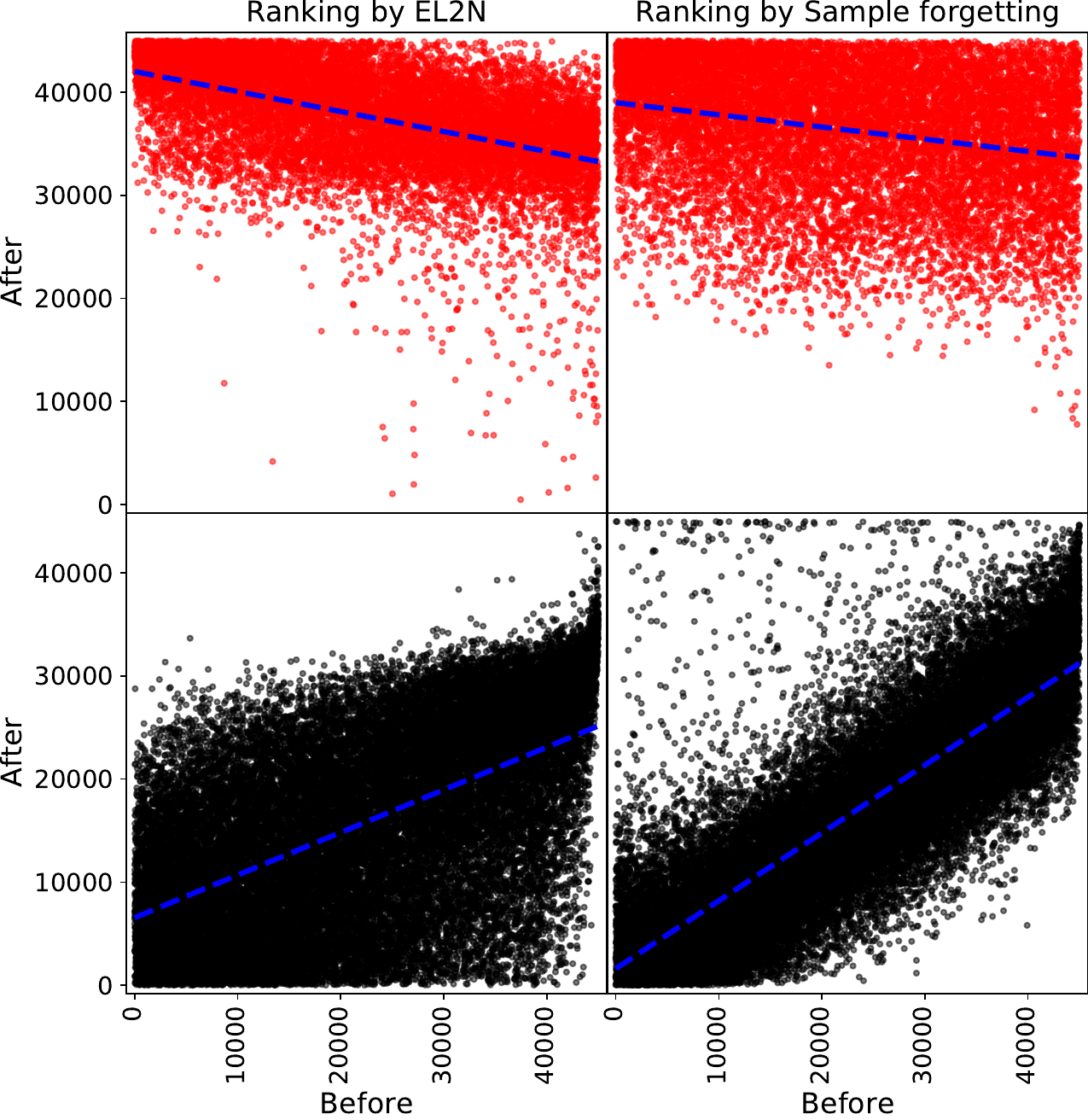}
    \caption{Sample ranking for synthetic (top-left), MNIST (top-right), and CIFAR-10 (bottom) before and after label noise is applied. Red samples have been label corrupted. Black samples were not corrupted. The blue dashed line is a linear best-fit line. The first column, in each plot, considers error $L_2$-norm (EL2N) scores. The second column considers sample forgetting scores.}
    \label{fig:synth_scores_before_after}
\end{figure}

% \begin{figure}
%     \centering
%     \includegraphics[width=\linewidth]{Figures/sampleforgetting_violin_plot.pdf}
%     \caption{Sample forgetting violin plot.}
%     \label{fig:sample_forgetting_violin}
% \end{figure}

% \begin{figure}
%     \centering
%     \includegraphics[width=\linewidth]{Figures/el2n_violin_plot.pdf}
%     \caption{EL2N violin plot.}
%     \label{fig:el2n_violin}
% \end{figure}





% LNCS Homepage, \url{http://www.springer.com/lncs}, last accessed 2023/10/25
% \end{thebibliography}

% \bibliographystyle{unsrt}
% \bibliography{refs}

\end{document}